\documentclass[runningheads]{llncs}
\usepackage{graphicx}
\usepackage{amsmath}
\usepackage{amssymb}
\usepackage{cite}
\usepackage{lipsum}
\usepackage{booktabs}
\usepackage{subcaption}
\begin{document}

%
\thispagestyle{empty}
{\noindent\Large Springer Copyright Notice}\\[1pt]

{\noindent Copyright (c) 2023 Springer

\noindent This work is subject to copyright. All rights are reserved by the Publisher, whether the whole or part of the material is concerned, specifically the rights of translation, reprinting, reuse of illustrations, recitation, broadcasting, reproduction on microfilms or in any other physical way, and transmission or information storage and retrieval, electronic adaptation, computer software, or by similar or dissimilar methodology now known or hereafter developed.}\\[0.5in]

{\noindent\large Accepted to be published in: Proceedings of the 12th Brazilian Conference on Intelligent Systems (BRACIS'23), Sep. 25--29, 2023.}\\[0.5in]

{\noindent Cite as:}\\[1pt]

{\setlength{\fboxrule}{0.5pt}
 \fbox{\parbox{0.95\textwidth}{J. M. Almeida, G. A. Castro, J. A. Machado-Neto and T. A. Almeida, ``An explainable model to support the decision about the  therapy protocol for AML,'' in \emph{Proceedings of the 12th Brazilian Conference on Intelligent Systems (BRACIS'23)}, Belo Horizonte, MG, Brazil, 2023, pp. 1--15.}}}\\[0.5in] 

{\noindent BibTeX:}\\[1pt]

{\setlength{\fboxrule}{1pt}
 \fbox{\parbox{1.2\textwidth}{
 @InProceedings\{BRACIS\_2023\_JMAlmeida,

 \begin{tabular}{p{2mm}lcl}
  & author    & = & \{J. M. \{Almeida\} and
               G. A. \{Castro\} and\\ & & &
               J. A. \{Machado-Neto\} and
               T. A. \{Almeida\}\},\\

  & title     & = & \{An explainable model to support the decision about the  therapy protocol\\
  &           &   & for AML\},\\

  & pages     & = & \{1--15\},\\

  & booktitle & = & \{Proc. of the 12th Brazilian Conference on Intelligent Systems (BRACIS'23)\},\\

  & address   & = & \{Belo Horizonte, MG, Brazil\},\\

  & month     & = & \{Sep. 11--15\},\\

  & year      & = & \{2023\},\\

  & publisher & = & \{\{Springer\}\},\\

  \end{tabular}

\}
 }}}

\clearpage

\title{An explainable model to support the decision about the  therapy protocol for AML\thanks{Supported by CAPES, CNPq, and FAPESP grant \#2021/13325-1.}}

\titlerunning{Explainable support decision about the therapy protocol for AML}

\author{Jade M. Almeida$^a$ \and
Giovanna A. Castro$^a$ \and
João A. Machado-Neto$^b$ \and
Tiago A. Almeida$^a$}

\authorrunning{J.M. Almeida et al.}

\institute{
$^1$Department of Computer Science (DComp-So)\\
Federal University of São Carlos (UFSCar)\\
18052-780, Sorocaba, São  Paulo -- Brazil\\
\vspace{0.25cm}
$^2$Institute of Biomedical Sciences\\
The University of S\~{a}o Paulo (USP)\\
05508-000, São  Paulo -- Brazil\\
\email{jade.almeida@dcomp.sor.ufscar.br, giovannacastro@estudante.ufscar.br, jamachadoneto@usp.br, talmeida@ufscar.br}}

\maketitle              

\baselineskip=0.936\baselineskip
\begin{abstract}
Acute Myeloid Leukemia (AML) is one of the most aggressive types of hematological neoplasm. To support the specialists' decision about the appropriate therapy, patients with AML receive a prognostic of outcomes according to their cytogenetic and molecular characteristics, often divided into three risk categories: favorable, intermediate, and adverse. However, the current risk classification has known problems, such as the heterogeneity between patients of the same risk group and no clear definition of the intermediate risk category. Moreover, as most patients with AML receive an intermediate-risk classification, specialists often demand other tests and analyses, leading to delayed treatment and worsening of the patient's clinical condition. This paper presents the data analysis and an explainable machine-learning model to support the decision about the most appropriate therapy protocol according to the patient's survival prediction.  In addition to the prediction model being explainable, the results obtained are promising and indicate that it is possible to use it to support the specialists' decisions safely. Most importantly, the findings offered in this study have the potential to open new avenues of research toward better treatments and prognostic markers.


\end{abstract}
\section{Introduction}
Acute Myeloid Leukemia (AML) is one of the most aggressive types of hematological neoplasm, characterized by the infiltration of cancer cells into the bone marrow. AML has decreasing remission rates regarding the patient's age, and its average overall survival rate is just 12 to 18 months~\cite{Pelcovits-2020}.

In 2010, the European LeukemiaNet (ELN) published recommendations for diagnosing and treating AML~\cite{Dohner-2010}, which became a field reference. A significant update to these recommendations was published in 2017~\cite{Dohner-2017} and 2022~\cite{Dohner-2022}, incorporating new findings concerning biomarkers and subtypes of the disease combined with a better understanding of the disease behavior.

For a diagnosis of AML, at least 10\% or 20\% of myeloblasts must be present in the bone marrow or peripheral blood, depending on the molecular subtype of the disease~\cite{Arber-2022}. This analysis is performed according to the Classification of Hematopoietic and Lymphoid Tissue Tumors, published and updated by the World Health Organization.

In addition to the diagnosis, the patient with AML receives a prognostic of outcomes, often divided into three risk categories: favorable, intermediate, and adverse. Cytogenetic and molecular characteristics define such stratification~\cite{Genomic-2013}. The cytogenetic characteristics come from certain chromosomal alterations. In turn, the molecular ones are determined according to mutations in the \textit{NPM1},  \textit{RUNX1}, \textit{ASXL1}, \textit{TP53}, \textit{BCOR}, \textit{EZH2}, \textit{SF3B1}, \textit{SRSF2}, \textit{STAG2}, and \textit{ZRSR2} genes. Specialists commonly use the ELN risk classification to support critical decisions about the course of each treatment, which can directly impact patients' quality of life and life expectancy. 

Patients with a favorable risk prognosis generally have a good response to chemotherapy. On the other hand, those with adverse risk tend not to respond well to this therapy, needing to resort to other treatments, such as hematopoietic stem cell transplantation~\cite{Genomic-2013}. The problem with the current risk prognosis is the high rate of heterogeneity between patients of the same risk group. In addition, there is no clear definition regarding the intermediate risk since these patients do not show a response pattern to treatments.

Most patients with AML receive an intermediate-risk classification~\cite{Dohner-2010}. Unfortunately, this makes specialists demand more information, such as the results of other tests and analyses, to support their decisions regarding the most appropriate treatment, even with little or no evidence of efficacy. This process can result in delayed initiation of treatment and consequent worsening of the patient's clinical condition.

To overcome this problem, this study presents the result of a careful analysis of real data composed of clinical and genetic attributes used to train an explainable machine-learning model to support the decision about the most appropriate therapy protocol for AML patients. The model is trained to identify the treatment guide that maximizes the patient's survival, leading to better outcomes and quality of life. 

\section{Related work}

The decision on therapy for patients with AML is strongly based on the prediction of response to treatment and clinical outcome, often defined by cytogenetic factors~\cite{Estey-2019}. However, the current risk classification can be quite different among patients within the same risk groups, in which the result can range from decease within a few days to an unexpected cure~\cite{Dohner-2010}.

Since the mid-1970s, the standard therapy for patients with AML has been chemotherapy, with a low survival rate. However, with advances, various data on mutations and gene expressions began to be collected, analyzed, and made available, accelerating the development of therapeutic practices.


In 2010, the European LeukemiaNet (ELN) proposed a risk categorization based on cytogenetic and molecular information, considering the severity of the disease~\cite{Dohner-2010}. This classification comprises four categories: favorable, intermediate I, intermediate II, and adverse. 

In 2017, a significant update to the ELN's risk classification was published~\cite{Dohner-2017}. The updated risk classification grouped patients into three categories (favorable, intermediate, and adverse) and refined the prognostic value of specific genetic mutations. Since then, specialists have commonly used this stratification to support important decisions about the course of each treatment, which can directly impact the patient's quality of life and life expectancy.

In 2022, the ELN's risk classification was updated again. The main change provided is related to the expression of the \textit{FLT3-ITD} gene. All patients with high expression but without any other characteristics of the adverse group are classified as intermediate risk. Another significant change is that mutations in \textit{BCOR}, \textit{EZH2}, \textit{SF3B1}, \textit{SRSF2}, \textit{STAG2}, and \textit{ZRSR2} genes are related to the adverse risk classification~\cite{Dohner-2022}. 

Specialists often rely on the ELN risk classification to define the treatment guidelines given to the patient shortly after diagnosis. Patients with a favorable risk generally present a positive response to chemotherapy. In contrast, patients with an adverse risk tend not to respond well to this therapy, requiring other treatments, such as hematopoietic stem cell transplantation~\cite{Genomic-2013}. However, there is no clear definition regarding the therapeutic response of AML patients with intermediate risk.

The problem with using the current risk classifications as a guide for deciding the most appropriate treatment is that there can be significant variability of patients in the same risk group, with different characteristics such as age and gender. For example, patients under 60 tend to respond better to high-dose chemotherapy. On the other hand, patients over 60 years old tend to have a low tolerance to intense chemotherapy and may need more palliative therapies~\cite{Lagunas-Rangel-2017}. Several studies suggest that age is a relevant factor when deciding the treatment for a patient, a fact that is not considered by the current risk classification. However, as most patients with AML receive the intermediate risk, specialists often require additional information, such as the results of other tests and analyses, to decide the most appropriate treatment, even with little or no evidence of efficacy~\cite{Dohner-2010}. This process can lead to a delay at the start of treatment and worsen the patient's clinical condition.

Studies have emphasized the significance of analyzing mutations and gene expression patterns in families of genes to determine the therapeutic course in AML. Over 200 genetic mutations have been identified as recurrent in AML patients through genomic research~\cite{Genomic-2013}. With genetic sequencing, the patient profile for AML has transitioned from cytogenetic to molecular~\cite{Ley-2008}. However, due to the heterogeneity of the disease, it is difficult to manually analyze the various genetic alterations that may impact the course of the disease. To overcome these challenges, recent studies have sought to apply machine learning (ML) techniques to automatically predict the outcome after exposure to specific treatments and complete remission of the disease.

For example, ~\cite{Ophir-2019} trained supervised ML models with data extracted from RNA sequencing and clinical information to predict complete remission in pediatric patients with AML. The $k$-NN technique obtained the best performance, with an area under the ROC curve equals to $0.81$. The authors also observed significant differences in the gene expressions of the patients concerning the pre-and post-treatment periods.

Later, \cite{Mosquera-2021} used clinical and genetic data to train a random forest classifier capable of automatically predicting the survival probability. According to the authors, the three most important variables for the model were patient age and gene expression of the \textit{KDM5B} and \textit{LAPTM4B} genes, respectively. The authors concluded that applying ML techniques with clinical and molecular data has great predictive potential, both for diagnosis and to support therapeutic decisions.

In the study of ~\cite{Gerstung-2017}, a statistical decision support model was built for predicting personalized treatment outcomes for AML patients using prognostic data available in a knowledge bank. The authors have found that clinical and demographic data, such as age and blood cell count, are highly influential for early death rates, including death in remission, which is mainly caused to treatment-related mortality. Using the knowledge bank-based model, the authors concluded that roughly one-third of the patients analyzed would have their treatment protocol changed when comparing the model's results with the ELN treatment recommendations. 

The success reported in these recent studies is an excellent indicator that recent ML techniques have the potential to automatically discover patterns in vast amounts of data that specialists can further use to support the personalization and recommendation of therapy protocols. However, one of the main concerns when applying machine learning in medicine is that the model can be explainable, and experts can clearly understand how the prediction is generated~\cite{Combi-2022}.

In this context, this study presents the result of a careful analysis of real data composed of clinical and genetic attributes used to train an explainable machine-learning model to support the decision about the most appropriate therapy protocol for AML patients. Our main objective is to significantly reduce the subjectivity involved in the decisions specialists must make and the time in the treatment decision processes. This can lead to robust recommendations with fewer adverse effects, increasing survival time and quality of life.

\section{Materials and methods\label{sec:materials-and-methods}}

This section details how the data were obtained, processed, analyzed, and selected. In addition, we also describe how the predictive models were trained.

\subsection{Datasets}\label{subsec:datasets}

The data used to train the prediction models come from studies by \textit{The Cancer Genome Atlas Program} (TCGA) and \textit{Oregon Health and Science University} (OHSU). These datasets are known as \textit{Acute Myeloid Leukemia}~\cite{Genomic-2013, Nature-2018} and comprise clinical and genetic data of AML patients. Both are real and available in the public domain at \url{https://www.cbioportal.org/}. We used three sets with data collected from the same patients: one with clinical information (CLIN), another with gene mutation data (MUT), and another with gene expression data (EXP). Table \ref{tab:summarize-data-begin} summarizes these original data.


\begin{table}[!htb]
    \centering
    \caption{Amount of original data in each database. Each database is composed of three sets of features: clinical information (CLIN), gene mutation data (MUT), and gene expression data (EXP)}
    \label{tab:summarize-data-begin}    
    \begin{tabular}{c|c|c|c|c|c}
        \cline{4-6}
        \multicolumn{3}{c}{}  & \multicolumn{3}{c}{Attributes} \\
        \toprule
             & Samples & Patients & CLIN & MUT & EXP \\
        \hline
        TCGA & 200     & 200      & 31     & 25,000    & 25,000      \\
        OHSU & 672     & 562      & 97     & 606      & 22,825       \\
        \bottomrule
    \end{tabular}
\end{table}

\subsection{Data cleaning and preprocessing}
\label{sec:preprocessing-data-cleaning}

Since the data comes from two sources, we have processed them to ensure consistency and integrity. With the support of specialists in the application domain, we removed the following spurious data:

\begin{enumerate}
    \item Samples not considered AML in adults observed by (\textit{i}) the age of the patient, which must not be less than 18 years, and (\textit{ii}) the percentage of blasts in the bone marrow, which should be greater or equal to 20\%;
    \item Samples without information on survival elapsed time after starting treatment (\textit{Overall Status Survival});
    \item Duplicate samples; and
    \item Features of patients in only one of the two databases.
\end{enumerate}

We used the 3-NN method to automatically fill empty values in clinical data features (CLIN). We used the features with empty values as the target attributes and filled them using the value predicted from the model trained with other attributes. Nevertheless, we removed the features of 37 genes with no mutations.



Subsequently, we kept only the samples in which all the variables are compatible, observing data related to the exams and treatment received by the patients, as these affect the nature of the clinical, mutation, and gene expression data. Of the 872 initial samples in the two databases, 272 were kept at the end of the preprocessing and data-cleaning processes. Of these, there are 100 samples from patients who remained alive after treatment and 172 who died before, during, or after treatment. Cytogenetic information was normalized and grouped by AML specialists. Moreover, the same specialists analyzed and grouped the treatments in the clinical data into four categories according to the intensity of each therapy:

\begin{enumerate}
\item \textit{Target therapy} -- therapy that uses a therapeutic target to inhibit some mutation/AML-related gene or protein;
\item \textit{Regular therapy} -- therapy with any classical chemotherapy;
\item \textit{Low-Intensity therapy} -- non-targeted palliative therapy, generally recommended for elderly patients; and
\item \textit{High-Intensity therapy} -- chemotherapy followed by autologous or allogenic hematopoietic stem cell transplantation.
\end{enumerate}

Finally, the specialists checked and validated all the data.

\subsection{Feature selection}\label{subsec:feature-selection}
This section describes how we have analyzed and selected the features used to represent clinical, gene mutation, and gene expression data.

\subsubsection{Clinical data}

\begin{figure}[!htb]
    \begin{center}
        \includegraphics[width=0.40\linewidth]{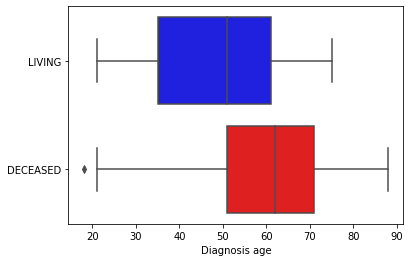}
        \includegraphics[width=0.40\linewidth]{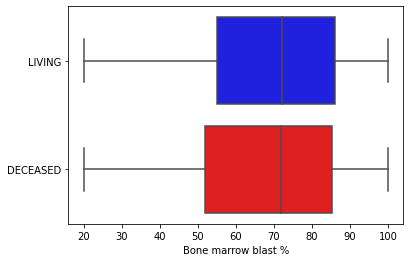}
        \includegraphics[width=0.40\linewidth]{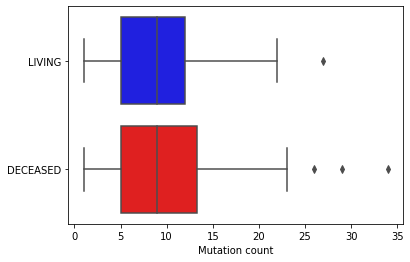}
        \includegraphics[width=0.40\linewidth]{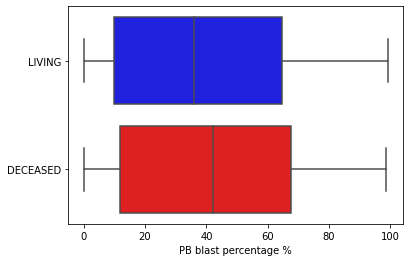}
        \includegraphics[width=0.40\linewidth]{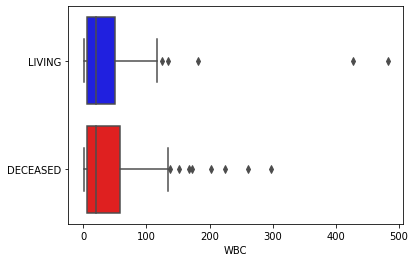}
    \end{center}
    \caption{Boxplots of continuous nature clinical features}
    \label{fig:boxplots-clinical}
\end{figure}

\begin{figure}[!htb]
    \begin{center}
        \includegraphics[width=0.40\linewidth]{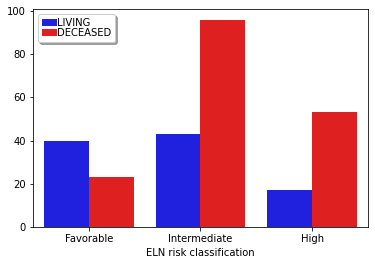}
        \includegraphics[width=0.4\linewidth]{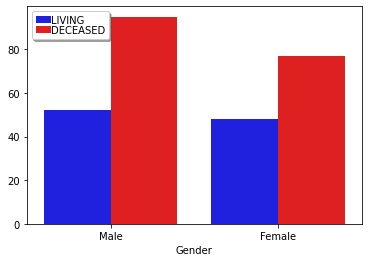}
        \includegraphics[width=0.40\linewidth]{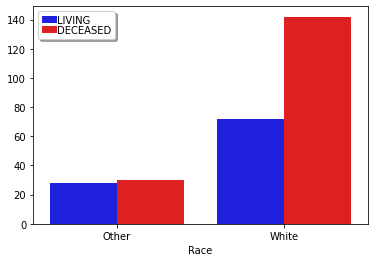}
        \includegraphics[width=0.40\linewidth]{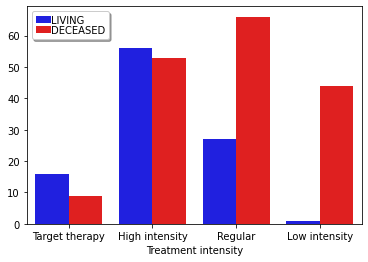}
    \end{center}
    \caption{Bar plots of categorical nature clinical features}
    \label{fig:barplots-clinical}
\end{figure}

Among the clinical attributes common in the two databases, specialists in the data domain selected the following 11 according to their relevance for predicting clinical outcomes. In Table~\ref{tab:feature-description}, we briefly describe all selected clinical features, and Table~\ref{tab:summarize-clinical-data} summarizes the main statistics of those with a continuous nature. Figures~\ref{fig:boxplots-clinical} and~\ref{fig:barplots-clinical} summarize their main statistics.

\begin{table}[!htb]
    \caption{Clinical features description}
    \label{tab:feature-description}
    \begin{tabular}{l|l}
        \toprule
        Feature & Description \\ \hline
        Diagnosis age & Patient age when diagnosed with AML \\
        \begin{tabular}[c]{@{}l@{}}Bone marrow  blast \%\end{tabular} & Percentage of blasts in the bone marrow \\
        Mutation count & Number of genetic mutations observed \\
        PB blast \% & Percentage of blasts in peripheral blood \\
        WBC & White blood cell count \\
        Gender & Patient gender \\
        Race & Whether the patient is white or not \\
        Cytogenetic info & \begin{tabular}[c]{@{}l@{}}Cytogenetic information the specialist used \\ in diagnosing the patient\end{tabular} \\
        \begin{tabular}[c]{@{}l@{}}ELN risk classification\end{tabular} & \begin{tabular}[c]{@{}l@{}}ELN risk groups\\  (favorable, intermediate, and adverse)\end{tabular} \\
        \begin{tabular}[c]{@{}l@{}}Treatment intensity\\  classification\end{tabular} & \begin{tabular}[c]{@{}l@{}}The intensity of treatment received by the patient\\  (target, regular, low-intensity,  or high-intensity therapy)\end{tabular} \\
        Overall survival status & Patient survival status (living or deceased). \\ 
        \bottomrule
    \end{tabular}
\end{table}

\begin{table}[!htb]
    \centering
    \caption{Main statistics of clinical features with a continuous nature}
    \label{tab:summarize-clinical-data}    
    \begin{tabular}{l|r|r|r|r}
        \toprule
        Feature & Minimum & Maximum & Median & Mean\\
        \hline
         Diagnosis age &  18 & 88 & 58 & 55.11\\
         Bone marrow blast \% & 20 & 100 & 72 & 68.13 \\
         Mutation count & 1 & 34 & 9 & 9.54\\
         PB blast \% & 0 & 99.20 & 39.10 & 40.99\\
         WBC & 0.4 & 483 & 39.40 & 19.85\\
         \bottomrule
    \end{tabular}
\end{table}

Among the clinical attributes, in line with several other studies, the only noticeable highlight is that the patient's age seems to be a good predictor of the outcome. The older the patient, the lower the chances of survival. All other attributes showed similar behavior for both classes, with subtle differences.

\subsubsection{Gene mutation data}
After cleaning and preprocessing the data, $281$ gene mutation features remained. Then, we employed the $\chi^2$ statistical method to select a subset of these features. We chose to use the $\chi^2$ test because it has been widely used in previous studies to analyze the correlation between genetic mutations and certain types of cancer~\cite{Rahman-2019}. We defined the following hypotheses: H0 -- patient survival is independent of gene mutation; and H1 -- both groups are dependent. Using $p < 0.05$, only two features were selected: \textit{PHF6} and \textit{TP53} gene mutations. 


The \textit{TP53} mutation is the best known among the two gene mutations selected. Several studies show the relationship between \textit{TP53} mutation with therapeutic response and prognosis. The \textit{TP53} gene is considered the guardian of genomic stability, as it controls cell cycle progression and apoptosis in situations of stress or DNA damage, and mutations in this gene are found in approximately half of the cancer patients~\cite{kastenhuber-2017-28886379}. Although mutations in \textit{TP53} are less common in AML patients (about 10\%), they predict a poor prognosis~\cite{grob-2022-35108372}.

The mutation in the \textit{PHF6} gene has been identified as a genetic alteration associated with hematologic malignancies\cite{Kurzer-2021}. \textit{PHF6} is a tumor suppressor gene, and several studies have shown a high mutation frequency in the adverse risk group of AML~\cite{Eisa-2023}. These observations suggest that \textit{PHF6} mutations may have a significant role in the development and progression of AML and may serve as a potential prognostic marker for the disease\cite{Van-Vlierberghe-2011}.

To further investigate the potential of gene mutation data on outcome prediction, we have enriched the set of gene mutation features with well-known genes already highlighted in studies in the literature~\cite{Lagunas-Rangel-2017, Pimenta-2021, Charrot-2020} and used by the ELN~\cite{Dohner-2022}. The literature features used were: \textit{FLT3}, \textit{NPM1}, \textit{DNMT3A}, \textit{IDH1}, \textit{IDH2}, \textit{TET2}, \textit{ASXL1}, \textit{RUNX1}, \textit{CEBPA}, \textit{NRAS}, \textit{KRAS}, \textit{SF3B1}, \textit{U2AF1}, \textit{SRSF2}.

\subsubsection{Gene expression data}

After data cleaning and preprocessing, 14,712 gene expression features remained. To select the most relevant features for outcome prediction, we have employed a method similar to Lasso Regression \cite{Tibshirani-1996}: we have trained an SVM model with L1 regularization. 
This method estimates the relevance of the features by assigning a weight coefficient to each of them. When a feature receives a zero coefficient, it is irrelevant enough for the problem the model was trained for. As a consequence, these features are not selected.


The method was trained with all 14,712 gene expression features, from which 22 were selected. 


The final datasets we have used to train and evaluate the outcome prediction models are publicly available at \url{https://github.com/jdmanzur/ml4aml\_databases}. It is composed of 272 samples (patient data) consisting of 11 clinical features (\texttt{CLIN}),  22 gene expression features (\texttt{EXP}), and  16 gene mutation features (\texttt{MUT}). Table~\ref{tab:final-data} summarizes each of these datasets.

\begin{table}[!htb]
    \caption{Final datasets used to train and evaluate the outcome prediction models}
    \label{tab:final-data}
    \begin{tabular}{l|c|l}
        \bottomrule
        Dataset & \#Features & Features \\ \hline
        \begin{tabular}[c]{@{}l@{}}Clinical  (\texttt{CLIN})\end{tabular} & 11 & 
        \begin{tabular}[c]{@{}l@{}}Diagnosis age, Bone marrow blast (\%), \\ Mutation count, PB blast (\%), \\ WBC, Gender, Race, \\ Cytogenetic info, ELN risk classification, \\ Treatment intensity classification, \\ Overall survival status (class)\end{tabular} \\
        \hline
        
        \begin{tabular}[c]{@{}l@{}}Gene expression (\texttt{EXP})\end{tabular} & 24 & 
        \begin{tabular}[c]{@{}l@{}} 
        \textit{CCDC144A}, \textit{CPNE8}, \textit{CYP2E1}, \\ \textit{CYTL1}, \textit{HAS1}, \textit{KIAA0141}, \\ \textit{KIAA1549}, \textit{LAMA2}, \textit{LTK}, \textit{MICALL2}, \\ \textit{MX1}, \textit{PPM1H}, \textit{PTH2R}, \textit{PTP4A3}, \\ \textit{RAD21}, \textit{RGS9BP}, \textit{SLC29A2}, \textit{TMED4},\\  \textit{TNFSF11}, \textit{TNK1}, \textit{TSKS}, \textit{XIST}\\
        Treatment intensity classification, \\ Overall survival status (class)
        \end{tabular} \\
        \hline
        \begin{tabular}[c]{@{}l@{}}Gene mutation (\texttt{MUT})\end{tabular} & 18 & 
        \begin{tabular}[c]{@{}l@{}}
        \textit{FLT3}, \textit{NPM1}, \textit{DNMT3A}, \textit{IDH1}, \textit{IDH2}, \\ 
        \textit{TET2}, \textit{ASXL1}, \textit{RUNX1}, \textit{CEBPA}, \textit{NRAS}, \textit{KRAS} \\ 
        \textit{SF3B1}, \textit{U2AF1}, \textit{SRSF2}, \textit{PHF6}, \textit{TP53},\\
        Treatment intensity classification, \\ Overall survival status (class)
        \end{tabular} \\ \bottomrule
    \end{tabular}
\end{table}

\subsection{Training the outcome prediction models}

Since interpretability is a crucial pre-requisite for machine-learning models in medicine~\cite{Combi-2022}, we have employed the well-known Explainable Boosting Machine (EBM) technique~\cite{Caruana-2015}.

EBM is a machine learning approach that combines the strengths of boosting techniques with the goal of interpretability. It is designed to create accurate and easily understandable models, making it particularly useful in domains where interpretability and transparency are important.

EBM extends the concept of boosting by incorporating a set of interpretable rules. Instead of using complex models like neural networks as weak learners, EBM employs a set of rules defined by individual input features. These rules are easily understandable and can be represented as ``if-then'' statements.

During training, EBM automatically learns the optimal rules and their associated weights to create an ensemble of rule-based models. The weights reflect the importance of each rule in the overall prediction, and the ensemble model combines their predictions to make a final prediction.

The interpretability of EBM comes from its ability to provide easily understandable explanations for its predictions. Using rule-based models, EBM can explicitly show which features and rules influenced the outcome, allowing AML specialists to understand the underlying decision-making process.

EBM has been applied successfully in various domains, such as predicting medical conditions, credit risk assessment, fraud detection, and predictive maintenance, where interpretability and transparency are paramount~\cite{pmlr-v139-nori21a}.

We have used the EBM classification method from the InterpretML library\footnote{InterpretML is a Python library that provides a set of tools and algorithms for interpreting and explaining machine learning models. The documentation is available at \url{https://interpret.ml/docs}.} to train seven outcome prediction models: one per dataset (CLIN, MUT, EXP) and four using all possible combinations (CLIN+MUT, CLIN+EXP, MUT+EXP, CLIN+MUT+EXP).

\subsection{Performance evaluation}

We evaluated the performance of the prediction models using holdout~\cite{Mitchell-1997}. For this, we have divided the data into three parts 80\% was randomly separated for training the models, 10\% of the remaining data was randomly selected for model and feature selection, and the remaining 10\% was used to test. The data separation was stratified; therefore, each partition preserves the class balance of the original datasets. We must highlight we performed the feature selection processes using only training and validation partitions.


We calculated the following well-known measures to assess and compare the performance obtained by the prediction models: accuracy, recall (or sensitivity), precision, F1-Score, and the Area Under the ROC Curve (AUC).

\section{Results and discussion}


First, we trained the outcome prediction models using only the best-known genes consolidated by studies in the literature, both for the expression and mutation contexts. These genes are \textit{FLT3}, \textit{NPM1}, \textit{DNMT3A}, \textit{IDH1}, \textit{IDH2}, \textit{TET2}, \textit{ASXL1}, \textit{RUNX1}, \textit{CEBPA}, \textit{NRAS}, \textit{KRAS}, \textit{SF3B1}, \textit{U2AF1}, and \textit{SRSF2}. Table \ref{tab:literature-results} presents the prediction performance obtained.

\begin{table}[!htb]
    \centering
    \caption{Results achieved by the outcome prediction models. The genes from MUT and EXP were selected according to consolidated studies in the literature}
    \label{tab:literature-results}    
        \begin{tabular}{llrrrrr}
        \toprule
        Model &  F1-Score &       AUC &  Accuracy &  Precision &    Recall \\
        \midrule
        CLIN                       &  0.57 &  0.53 &  0.57 &   0.57 &  0.57 \\
        MUT                       &  0.64 &  \textbf{0.70} &  0.64 &   \textbf{0.76} &  0.64 \\
        EXP                        &  0.66 &  0.62 &  \textbf{0.68} &   0.66 &  \textbf{0.68} \\
        CLIN+MUT           &  \textbf{0.67} &  0.64 &  \textbf{0.68} &   0.67 &  \textbf{0.68} \\
        CLIN+EXP            &  0.57 &  0.53 &  0.57 &   0.57 &  0.57 \\
        MUT+EXP             &  0.63 &  0.59 &  0.64 &   0.63 &  0.64 \\
        CLIN+MUT+EXP &  0.54 &  0.51 &  0.54 &   0.55 &  0.54 \\
        \bottomrule
        \end{tabular}
\end{table}

The model that achieved the best result was the one that combined clinical and genetic mutation data. When analyzing the models trained with individual datasets, the ones based on gene mutation and expression showed the best performances. However, the overall results obtained are low and unsatisfactory for predicting the outcomes of AML patients. Surprisingly, the genes most known in the literature seem not strongly associated with outcomes prediction.


We then trained the outcome prediction models using the data resulting from the pre-processing, data analysis, and feature selection process described in Section~\ref{sec:materials-and-methods} (Table~\ref{tab:final-data}). Table \ref{tab:all-results} shows the results obtained.

\begin{table}[htb!]
    \centering
    \caption{Results achieved by the outcome prediction models. The genes from MUT were selected using $\chi^2$-test + the genes selected according to the literature.  The genes from EXP were selected using LASSO}
    \label{tab:all-results}    
    \begin{tabular}{llrrrrr}
        \toprule
        Model &  F1-Score & AUC &  Accuracy &  Precision &    Recall \\
        \midrule
        CLIN                        &  0.57 &  0.53 &  0.57 &   0.57 &  0.57 \\
        MUT                         &  0.65 &  0.63 &  0.64 &   0.66 &  0.64 \\
        EXP                           &  \textbf{0.86} &  \textbf{0.84} &  \textbf{0.86} &   \textbf{0.86} &  \textbf{0.86} \\
        CLIN+MUT              &  0.67 &  0.64 &  0.68 &   0.67 &  0.68 \\
        CLIN+EXP                &  0.78 &  0.74 &  0.79 &   0.78 &  0.79 \\
        MUT+EXP                &  0.82 &  0.79 &  0.82 &   0.82 &  0.82 \\
        CLIN+MUT+EXP     &  0.78 &  0.74 &  0.79 &   0.78 &  0.79 \\
        \bottomrule
    \end{tabular}
\end{table}


The performance of the model trained only with the mutation data deteriorated slightly compared to the one obtained only with the genes highlighted in the literature. However, the performance of the model trained only with the expression data showed a remarkable improvement since all performance measures were up about 30\%, and figuring as the best model we achieved. This strong increase in model performance is probably due to the careful KDD (Knowledge Discovery in Databases) process performed on the data and the new genes discovered to be good predictors.

Since gene expression data are expensive to obtain, they are usually absent on the first visit with specialists~\cite{Dohner-2017}. In this case, the outcome prediction model trained with clinical data and genetic mutations can be used as an initial guide to support the first therapeutic decisions.

\begin{figure}[!htb]
    \begin{center}
        \includegraphics[width=0.9\linewidth]{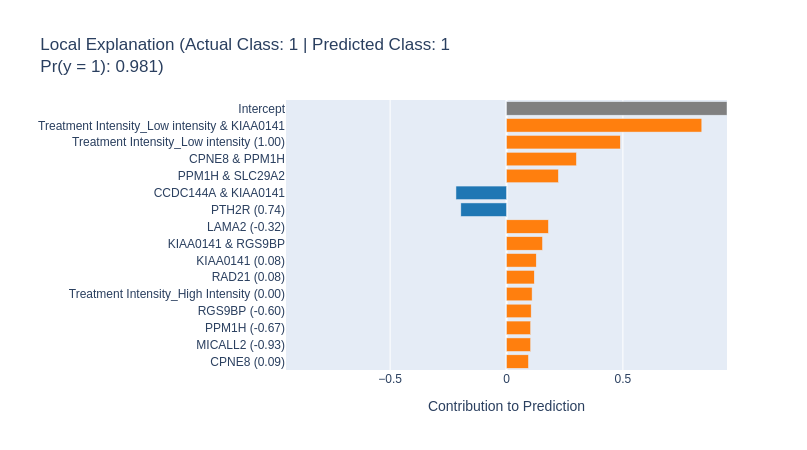}
        \includegraphics[width=0.9\linewidth]{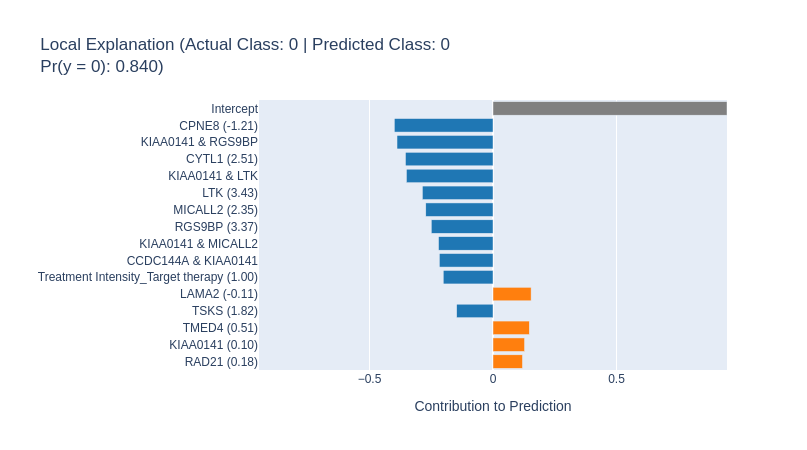}
    \end{center}
    \caption{Local explanation showing how much each feature contributed to the prediction for a single sample using the classification model trained with the EXP feature set. The intercept reflects the average case presented as a $log$ of the base rate (e.g., $-2.3$ if the base rate is $10\%$). The 15 most important terms are shown.}
    \label{fig:prediction-interpretability}
\end{figure}

The main advantage of using EBMs is that they are highly intelligible because the contribution of each feature to an output prediction can be easily visualized and understood. Since EBM is an additive model, each feature contributes to predictions in a modular way that makes it easy to reason about the contribution of each feature to the prediction. Figure~\ref{fig:prediction-interpretability} shows the local explanation for two test samples correctly classified as positive and negative using the classification model trained with the EXP feature set.

\begin{figure}[!htbp]
    \centering
    \begin{subfigure}{0.49\textwidth}
         \includegraphics[width=\textwidth]{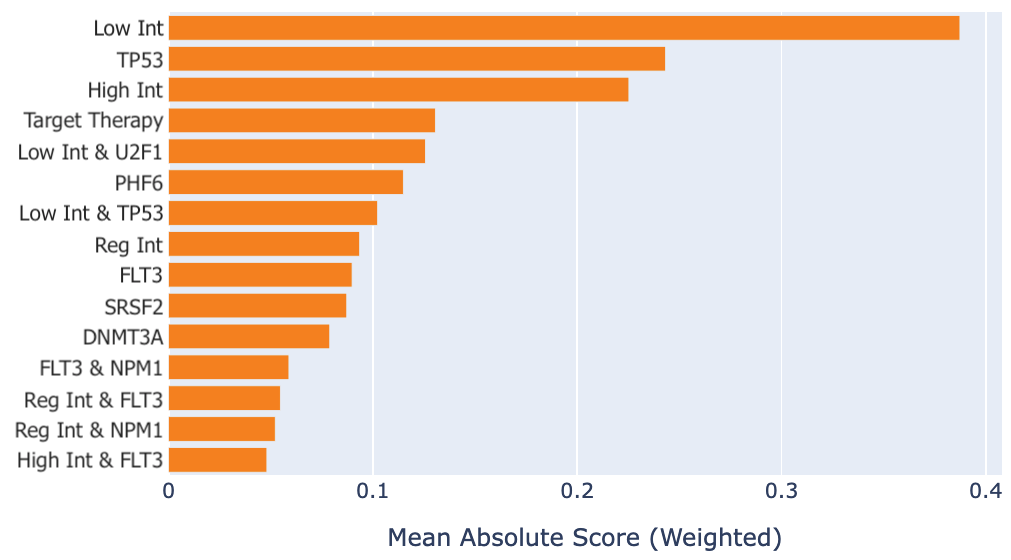}
         \caption{Gene mutation data}
         \label{fig:feat_imp_MUT}
     \end{subfigure}
     \hfill
     \begin{subfigure}{0.49\textwidth}
         \includegraphics[width=\textwidth]{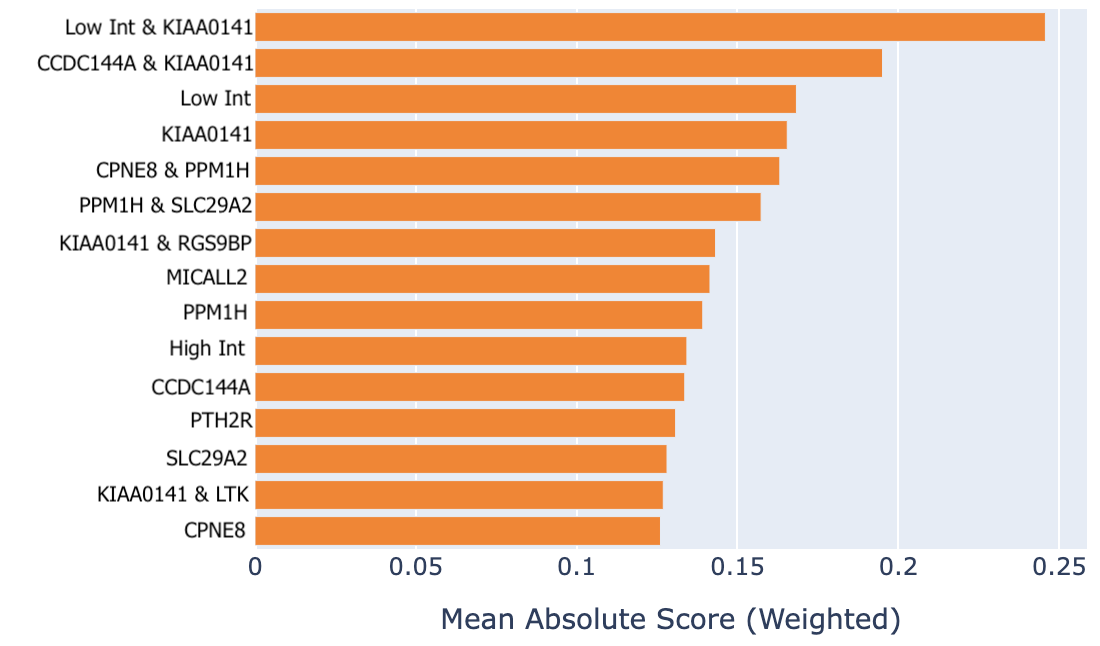}
         \caption{Gene expression data}
         \label{fig:feat_imp_EXP}
     \end{subfigure}
     \hfill    
    \begin{subfigure}{0.5\textwidth}
         \includegraphics[width=\textwidth]{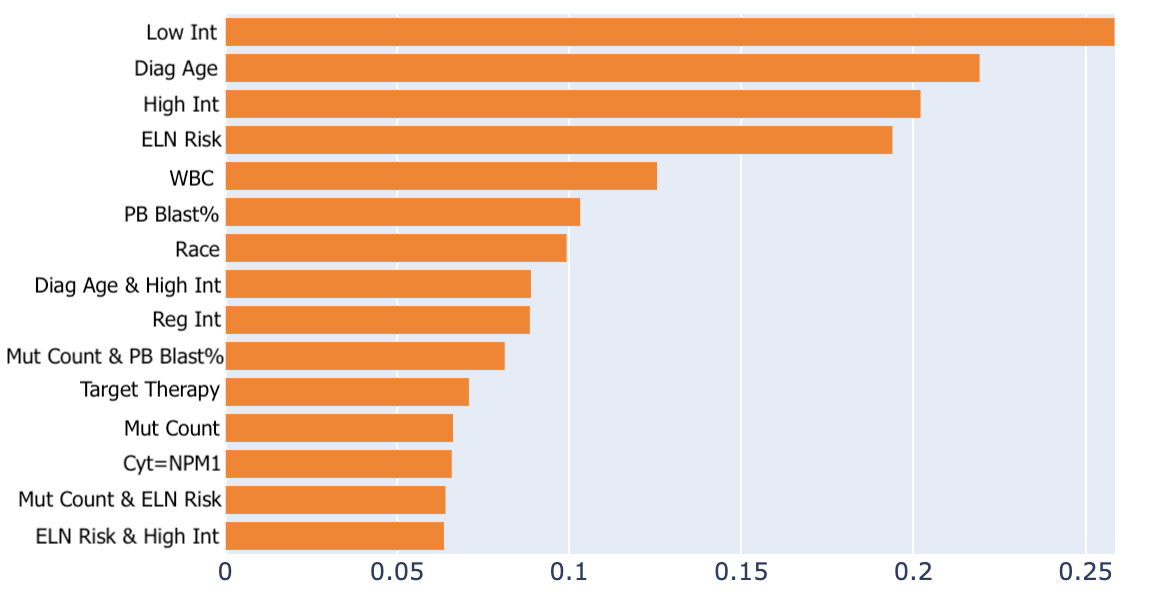}
         \caption{Clinical data}
         \label{fig:feat_imp_CLIN}
     \end{subfigure}
    
    \caption{Top-15 features that most influence the models' prediction}
    \label{fig:feature-importance}
    \end{figure}

Figure~\ref{fig:feature-importance} presents the top-15 attributes according to their importance in generating the prediction of outcome using gene mutation (Fig~\ref{fig:feat_imp_MUT}), gene expression (Fig~\ref{fig:feat_imp_EXP}), and clinical data (Fig~\ref{fig:feat_imp_CLIN}), respectively. The attribute importance scores represent the average absolute contribution of each feature or interaction to the predictions, considering the entire training set. These contributions are weighted based on the number of samples within each group.

The four most influential clinical features are (\textit{i}) when low-intensity treatment is chosen by the specialist; (\textit{ii}) the patient's age; (\textit{iii}) when high-intensity treatment is chosen; and (\textit{iv}) the ELN risk classification. It is well-known that the age at diagnosis and the ELN risk classification can potentially impact the patient's outcome \cite{Dohner-2022, Arber-2022}. Considering that specialists often do not have access to the most suitable treatment intensity during model prediction, the predictions are automatically generated for the four categorized treatment types (Section \ref{subsec:datasets}), and the one that best optimizes the patient's survival time is selected as the recommended therapy.



Regarding genetic mutation data, the mutations in the \textit{TP53} and \textit{PHF6} genes are ranked as the most influential, followed by the gene mutations already well-known in the literature. If, on the one hand, the mutation in the \textit{TP53} gene was already expected, to the best of our knowledge, there are no studies in the literature associating the \textit{PHF6} gene with predicting outcomes in the context of AML. Therefore, laboratory tests should be performed to confirm whether this gene may serve as a potential prognostic marker.




Among the most influential genetic expression features for model prediction, the following stand out \textit{KIAA0141}, \textit{MICALL2}, and \textit{SLC9A2}. Unlike the other genes, such as \textit{PPM1} and \textit{LTK}, which are already related in several AML studies, as far as we know, there is no study in the literature relating any of the three genes mentioned in the context of AML. In particular, the gene \textit{KIAA0141}, also known as \textit{DELE1}, has been recently identified as a key player~\cite{Sharon-2023}. 
In a pan-cancer analysis, \textit{MICALL2} was highly expressed in 16 out of 33 cancers compared to normal tissues~\cite{lin-2022-35281853}.  The role of \textit{SLC9A2} in cancer is still an area of active research, and the exact relationship between \textit{SLC9A2} and cancer development or progression is not fully understood. However, some studies have suggested potential associations between \textit{SLC9A2} and certain types of cancer, such as colorectal, breast, and gastric cancer.

The findings presented in this paper suggest that the biological role of these genes in the pathogenesis and progression of AML deserves future functional studies in experimental models and may provide insights into the prognosis and the development of new treatments for the disease.

\section{Conclusion}

To support the decision on the therapy protocol for a given AML patient, specialists usually resort to a prognostic of outcomes according to the prediction of response to treatment and clinical outcome. The current ELN risk stratification is divided into favorable, intermediate, and adverse. Despite being widely used, it is very conservative since most patients receive an intermediate risk classification. Consequently, specialists must require new exams, delaying treatment and possibly worsening the patient's clinical condition.


This study presented a careful data analysis and explainable machine-learning models trained using the well-known Explainable Boosting Machine technique. According to the patient's outcome prediction, these models can support the decision about the most appropriate therapy protocol.  In addition to the prediction models being explainable, the results obtained are promising and indicate that it is possible to use them to support the specialists' decisions safely. 

We showed that the prediction model trained with gene expression data performed best. In addition, the results indicated that using a set of genetic features hitherto unknown in the AML literature significantly increased the prediction model's performance. The finding of these genes has the potential to open new avenues of research toward better treatments and prognostic markers for AML.

For future work, we suggest collecting more data to keep the models updated regarding the disease variations over time. Furthermore, the biological role of the genes \textit{KIAA0141}, \textit{MICALL2}, \textit{PHF6}, and \textit{SLC92A} in the pathogenesis and progression of AML deserves functional studies in experimental models.

%
%
\bibliographystyle{splncs04}

\end{document}